\documentclass[twocolumn, 8pt]{article}

\usepackage{JAMIA}
\usepackage{adjustbox}
\usepackage{booktabs}

\title{Design and Evaluation of Epira: An Automated Interview System for Case Investigation and Contact Tracing
}
\author[1]{Ishita Asnani\textsuperscript{*}}
\author[2]{Shaghayegh Fazliani, MSc\textsuperscript{*}}
\author[2]{Zachary Ryan}
\author[2]{Kriyoshig Prasai}
\author[2]{Chirag Kumar, BA}
\author[2]{Indra Gonzalez Ojeda, PhD}
\author[2]{Marc Lipsitch, DPhil}
\author[2]{Madeleine Udell, PhD}
\affil[1]{University of Illinois Urbana-Champaign, Urbana, IL, USA}
\affil[2]{Stanford University, Stanford, CA, USA}
\date{}

\begin{document}

\par\noindent\rule[-7pt]{15.5cm}{0.2em}
\begin{strip}
    \begin{minipage}{.88\textwidth}
        \maketitle
        \vspace{1em}
        {\jamiasans
        \textbf{\textit{* Ishita Asnani and Shaghayegh Fazliani contributed equally to this article.}}\\[1em]
        \textbf{Corresponding Author:} Shaghayegh Fazliani; ORCID: 0009-0000-3763-2261; Mailing address: 450 Jane Stanford Way, Building 380, Department of Mathematics, Stanford, CA, USA 94305-2125; Email: fazliani@stanford.edu\\[0.3em]

        \small
        \abstractSection
        {Case investigation and contact tracing are labor-intensive. Web forms struggle when respondents need clarification before disclosing contacts, and open-ended chatbots may deviate from protocols. We developed and evaluated Epira, a protocol-controlled large language model (LLM) interviewer for multi-disease case investigation and contact tracing.} % Objective
        {Epira pairs LLM dialogue with a rule-based controller enforcing required questions, safety handoffs, and a structured case record, and adapts to new diseases by questionnaire upload. Local (Qwen2.5) and cloud (GPT-5.4) backends were tested. Study 1 used 108 simulated cases with ground-truth fields and contacts for measles and gonorrhea under Epira and open-ended conditions (864 sessions). Study 2 enrolled 72 volunteers in a 2$\times$2 disease-by-model interview assessing completion, usability, acceptability, and trust.} % Materials and Methods
        {Across Study 1, Epira with a local LLM captured 88.9\% of required fields and 60.1\% of ground-truth contacts. Open-ended interviewers recovered 31.1--41.9 percentage points fewer required fields than the same models in Epira. In Study 2, completion was 100\%; mean adapted System Usability Scale (SUS) was 82.95 $\pm$ 16.17 on a 0–100 scale and mean trust was 3.85 $\pm$ 0.77 on a 1–5 scale.} % Results
        {Protocol control drove completeness more than model choice. Local and cloud deployments produced similar records, supporting privacy-preserving local use with human review of routed interviews and missed contacts.} % Discussion
        {Epira recovers required fields and contacts under diverse disclosure conditions, works with local and cloud LLMs, and received favorable usability and trust ratings.} % Conclusion
        {automated interviewer; public health; contact tracing; case investigation; large language model (LLM)} % Key words
}
        \par\noindent\rule[-7pt]{15.5cm}{0.2em}
        \hspace{2cm}
    \end{minipage}
\end{strip}

\section*{Introduction}

Case investigation and contact tracing are essential public-health responses to infectious disease, but they remain time-consuming and difficult to scale.\cite{hellewell2020,lash2021} During COVID-19, health departments faced large caseloads and limited interviewer capacity.\cite{lash2021} Even outside of pandemic periods, health departments may be confronted with costly surges of case investigations or contact tracing needs.\cite{marx2017} Many health departments lack the personnel to systematically investigate cases of certain infections,\cite{sriudomporn2026} a constraint that has grown as budgets have tightened. Automated interviewing tools that are efficient and usable across diseases can expand public health investigation capacity.

Existing digital approaches leave an important gap. Web forms can collect structured data, but they fail when respondents have limited literacy, are hesitant to disclose contacts, or need conversational clarification. Fully open-ended artificial intelligence chatbots can sound natural, yet they may skip required questions, give unsafe advice, invent missing details, or wander away from the approved interview protocol.\cite{huo2025}

\begin{figure*}[!htbp]
    \centering
    \setlength{\abovecaptionskip}{0pt}
    \includegraphics[width=0.8\textwidth]{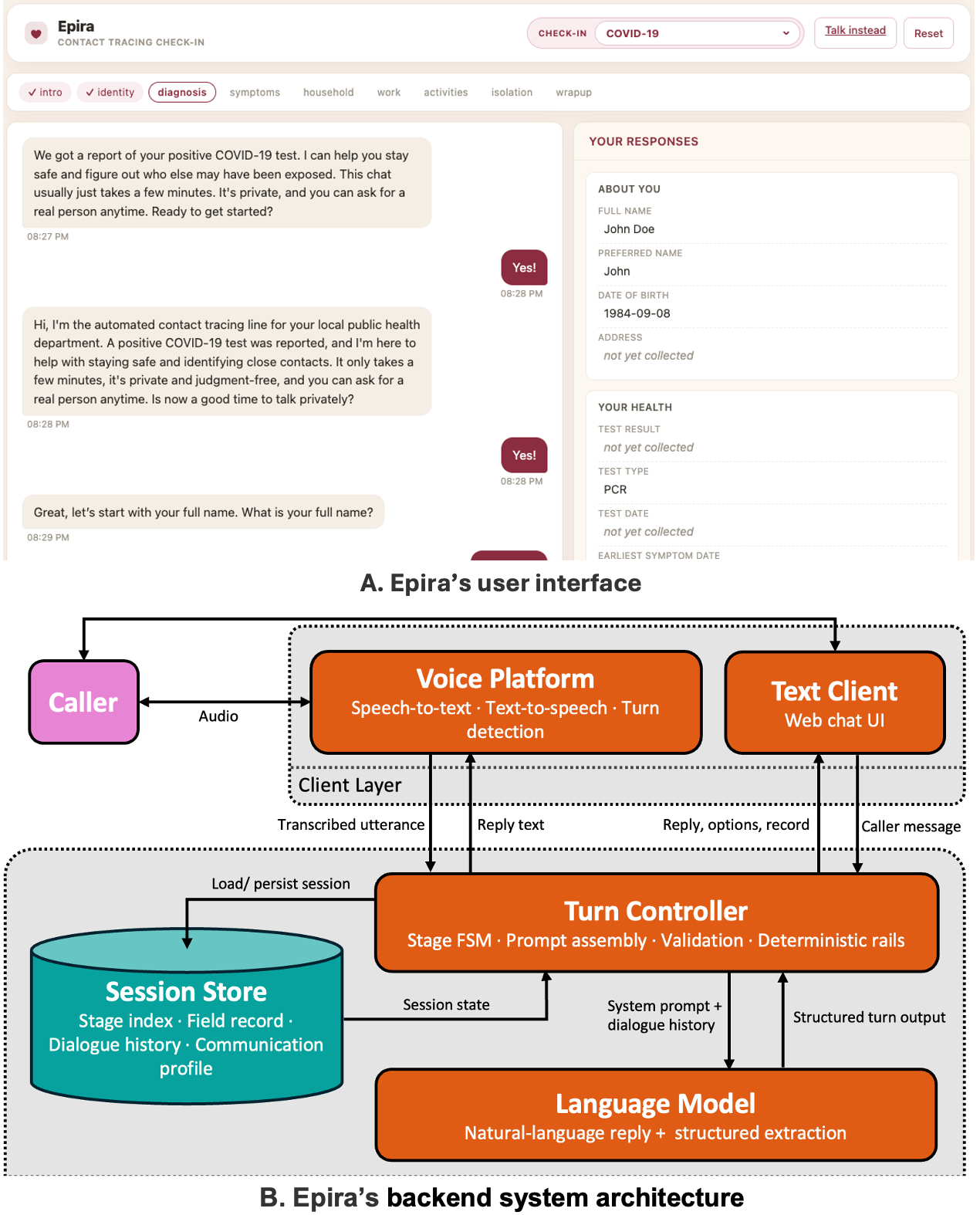}
    \caption{\textbf{Epira's System.}
    \textbf{(A) Epira's user interface.} The page shows the dialogue on the left and a structured case form on the right. The form fills in as the interview proceeds, from the answers Epira has accepted, and the respondent can edit any field at any time to correct it. A completion bar at the top of the page shows which stage the interview is in and how many stages remain. A Talk instead button at the top right switches the same interview from text to a voice call. \textbf{(B) Epira's backend system architecture}. One controller runs every disease. The active disease is a specification (stages, required fields, and safety rules), so adding a disease does not change the model or the controller. On each turn the caller's message, typed in the web chat or spoken through the voice service, is passed to the controller with the current stage prompt, the structured case record, and recent dialogue. A language model drafts the next reply and proposes structured updates and safety flags. Protocol rules then accept, correct, or reject that proposal. Only the accepted reply is returned to the caller, and only accepted fields are written to the case record. The voice path is part of the system shown here. Study 1 and Study 2 evaluated the text interface only.}
    \label{fig:epira_system}
\end{figure*}

Public-health practice needs interviewing tools that are both conversational and accountable. This paper describes Epira, an automated system for case investigation and contact tracing that couples natural dialogue with protocol control and a live structured case record (Figure~\ref{fig:epira_system}A). The same interface supports text or voice and adapts the interview engine to new diseases by uploading new disease protocols. Epira currently supports contact tracing for COVID-19, measles, and gonorrhea, travel-intake monitoring for Ebola, and exposure investigation for Cyclospora.

We developed Epira and evaluated whether it can recover required case fields and contacts under diverse disclosure conditions. In simulated interviews with known ground truth, we compared Epira with a local open-source model (Qwen2.5) and a cloud model (GPT-5.4) as the interview engine (Table~\ref{tab:study1_models}), and against the same models used as open-ended chatbots without protocol control (Table~\ref{tab:study1_openended}). A human volunteer pilot assessed completion, usability, acceptability, and trust. (Table~\ref{tab:study2}).

% \noindent{\footnotesize\textit{Alt text:} Two-panel figure of Epira. Panel A shows the web chat interview UI with a live structured case form. Panel B is a flowchart of the client layer, turn controller, session store, and language model.}
% \FloatBarrier

\section*{Methods}

\subsection*{System Design and Procedures}
Epira conducts case investigation and contact tracing one conversational turn at a time. The respondent's message, typed or transcribed from speech, is sent to an LLM with the conversational context (described in the next paragraph). The model returns a JSON object with three parts: the next utterance, candidate updates to the case record, and safety flags. A deterministic controller (``interview controller''; see Figure~\ref{fig:epira_system}B) then accepts, revises, or rejects each part. Only the accepted utterance is shown, and only accepted fields are written. Otherwise, the model rephrases the question to elicit a usable answer. The interface displays the dialogue next to a live case form that the respondent can edit (Figure~\ref{fig:epira_system}A).

The conversational context for each LLM call consists of: 1) The fields still required in the current interview stage and the safety rules for that stage. 2) The structured case record, including every field and contact already stored. 3) The dialogue history, including every turn in the current stage and the last eight turns from earlier stages. Stored fields remain in the final case record even after they leave this window. The smaller local model restricts context by recording only the field names defined in the current stage.

Interview progress is governed by a disease-specific stage sequence. Communicable (contagious) disease protocols, which emphasize contact elicitation, share the same interview stage sequence: (1) introduction and consent to continue; (2) identity verification (name and date of birth); (3) diagnosis and testing history; (4) symptoms and emergency warning signs; (5) household members and home contacts; (6) workplace setting and work contacts; (7) recent activities, places visited, and other close contacts; (8) isolation or monitoring guidance; and (9) wrap-up and closing. Protocols for noncommunicable infections (for example, foodborne illness such as cyclosporiasis) follow the same introduction, identity, diagnosis, and symptom stages. The activities and exposures stage expands to capture relevant risk factors, such as foods eaten, water sources, travel, and related exposure details, before wrap-up.

Before a reply is shown to the user, fixed rules check the model's proposal against the current stage. A proposed answer is written to the case record only if it belongs to a field allowed in that stage. The stage does not advance until its required fields are filled or explicitly recorded as none. If the respondent reports emergency symptoms or asks for a person, the automated interview stops and the session is routed for human follow-up. If the same question is repeated without new information, an anti-stall rule skips the field and forces the interview forward. Implementation details of the rules and handoff criteria appear in Appendix Section 2. These deterministic checks are what keep the interview inside the approved protocol even when the language model proposes an incomplete or out-of-stage update.

One engine runs every disease without changing the model or the controller. To add a new disease to Epira, an administrator enters a questionnaire, or asks Epira to draft one from a list of trusted public-health sources (CDC, CSTE, WHO, CIFOR, and U.S. state health departments), with a source citation on every drafted question. The system turns that list into an interview specification: the disease-specific questions form the middle stages, and a fixed frame is added automatically (introduction and consent, identity, wrap-up, and the emergency, human-handoff, and decline exits). The specification file lists stages, fields, answer options, which fields are required, stage transitions, and safety routing. A certified reviewer can then edit each question, its answer options, and whether it is required. Epira enables the interview protocol for the new disease when the reviewer marks it for deployment. The trusted source list, question format, review steps, and a figure demonstrating system adaptation to a new disease appear in Supplementary Appendix Section 1.

\subsection*{Study Design, Datasets, and Participants}
We evaluated Epira in two complementary studies: a paired simulation benchmarking study (Study 1) and a non-randomized four-arm, single-session usability pilot (Study 2). Both use the measles and gonorrhea interview specifications. Study 1 assessed interview performance in simulated index cases with known ground truth for required fields and contacts. Study 2 crossed disease (measles or gonorrhea) with interviewer backend (local Qwen2.5 or cloud GPT-5.4). Human volunteers role-played fictional cases, and the pilot evaluated usability and acceptability rather than clinical or health outcomes. Study implementation details appear in Supplementary Appendix Sections 4 and 5.

In Study 1, each simulated index case had a hidden answer key for required case fields and a hidden true contact list. Scoring was based on this ground truth. We used two case banks. The routine bank comprised 36 cases (190 known contacts for measles and 65 for gonorrhea) written to resemble typical cooperative interviews. The challenging bank comprised 72 cases (339 known contacts for measles and 128 for gonorrhea) written to include resistant disclosure, sparse or withheld contacts, and scenarios that should trigger human handoff. Banks were analyzed separately. Simulated respondents were generated by a fixed language model (GPT-5.4-mini) held constant across all studies. All interviews used Epira's text interface in English. As a baseline, we also asked an open-ended chatbot to interview the same cases, with minimal instructions and no protocol rules, and evaluated the same metrics (Table~\ref{tab:study1_openended}). We ran each experiment with both Qwen2.5 and GPT-5.4 backends to assess tradeoffs. The 108 unique personas were each evaluated for both diseases, with both interviewer models and under both structuring conditions (Epira and open-ended), yielding 864 interview sessions. Holding the simulated respondent model fixed across conditions isolates differences due to the interviewer protocol and backend rather than variation in how cases answered.

In Study 2, we recruited English-speaking adults aged 18 years or older and located in the United States through Prolific. Participants were recruited into four separate Prolific study conditions, each linked to one prespecified disease--model combination. Participants reviewed a study information sheet and agreed to participate. They completed one text interview in a 2$\times$2 design (measles or gonorrhea $\times$ local Qwen2.5 or cloud GPT-5.4). Eighteen participants were enrolled in each arm (72 total). Each participant was assigned one case from the routine bank and role-played that fictional person. Fictional scenarios were assigned sequentially in a prespecified order by the study server, with repeat requests from the same Prolific participant ID returning the same scenario. They were instructed to use only the assigned scenario and not to enter real health or identifying information. No audio or video was recorded. Participants were not told which model backend powered the interview. Investigators and analysts were not masked to study condition. After the interview, each participant completed a questionnaire on acceptability and usability (Table~\ref{tab:study2}). The session was described as about 20 minutes, and participants were paid \$5 through Prolific. No names or direct contact information were collected in the research dataset. Sex assigned at birth, gender identity, race, ethnicity, and disability status were not collected because this pilot was designed to estimate overall usability rather than demographic subgroup differences. Participation required English proficiency, being based in the United States, internet access, and use of an online text-based interface. Disability-specific accessibility was not evaluated at this stage.

\subsection*{Performance Metrics and Outcomes}
\textbf{Study 1:} Primary outcomes were pre-specified and scored from the structured interview record against known ground truth (Tables~\ref{tab:study1_models}--\ref{tab:study1_openended}; Figure~\ref{fig:study_stats}A). Field completeness was the proportion of required case fields captured. Contact completeness was the pooled proportion of ground-truth contacts matched: total matched contacts divided by total ground-truth contacts (personas with more true contacts contribute more to the pooled ratio). Interview efficiency was information gained per interview turn---information units captured by the end of the interview divided by the number of turns, where one ``information unit'' is one required non-contact case field filled or one distinct contact recorded, and one ``turn'' is one interviewer-respondent exchange. Average number of interview turns was the mean number of turns per interview. We also reported LLM API cost per completed interview. Secondary outcomes included contact accuracy and fabrication, performance by respondent type, and correctness of emergency and human-handoff routing. Formal definitions and required fields by disease and stage appear in Supplementary Appendix Sections 3 and 4.4.

\textbf{Study 2:} Outcomes came from interview completion and a required post-interview questionnaire (Table~\ref{tab:study2}; Figure~\ref{fig:study_stats}B). Questionnaire items used a 1 (strongly disagree) to 5 (strongly agree) scale. Interview completion was the percentage of enrolled participants in each arm who finished the chat. Usability was measured with an adapted ten-item System Usability Scale (SUS).\cite{brooke1996} Items kept the standard SUS structure and scoring but substituted ``interview tool'' for the original product wording. Responses were transformed to a 0--100 score (higher = more usable).

The remaining items on the survey used a 1--5 scale (1 = strongly disagree, 3 = neutral, 5 = strongly agree). Three items assessed trust and comfort; whether the tool seemed to attend carefully to answers, comfort answering questions, and privacy worry (reverse-scored so higher = greater privacy comfort). Their mean was the primary trust index. Preferences and willingness to complete a similar interview and to name contacts if newly positive (reported as their mean), preference for the tool over a long online form, and preference for a human interviewer over the tool were also assessed. We also assessed interview burden and problems; whether the interview felt too long, wanting to stop for a human, and encountering technical problems (all 1--5; higher = greater burden or difficulty).

\subsection*{Statistical analysis}
In Study 1, each simulated case was interviewed once under each condition, and the two results for that case were kept together. In Table~\ref{tab:study1_models}, the difference is the local model (Qwen2.5) minus the cloud model (GPT-5.4). In Table~\ref{tab:study1_openended}, the difference is Epira minus the open-ended interviewer, calculated separately for each model. Differences in percentages are in percentage points. The 95\% confidence interval came from repeating that calculation on 10,000 resamples of the cases. Because each case was run once, the interval reflects variation from case to case. In Study 2, analyses were conducted separately for the four prespecified disease--model conditions. Questionnaire outcomes are reported as mean $\pm$ SD. The questionnaire required a response to every scored item; therefore, no item-level values were imputed. Between-arm analyses were descriptive because the study was not randomized or powered for confirmatory comparisons. The single usability and trust values in the abstract are the unweighted means of the four arm means. Statistical analyses were implemented in Python scripts using NumPy version 2.4.4 and SciPy version 1.17.1. The resampling procedure is described in Supplementary Appendix Section 6.

\subsection*{Ethics}
Study 1 used synthetic cases and did not involve human subjects. Study 2 was reviewed by the Stanford University Institutional Review Board and determined to be exempt (eProtocol 88288; Exempt Category 3). Participants received a study information sheet and a consent page and agreed to participate.

\section*{Results}

\subsection*{Study 1}
Epira was compared with a local model (Qwen2.5) and a cloud model (GPT-5.4) on measles and gonorrhea, separately for the routine bank (36 cases) and the challenging bank (72 cases) (Table~\ref{tab:study1_models}). Differences are local minus cloud; positive values favor the local model.

\begin{table*}[!htbp]
\setlength{\abovecaptionskip}{5pt}
\caption{Primary outcomes of the simulated evaluation of Epira for measles and gonorrhea (Study 1) with two LLMs as the interview engines.
Each block reports one disease and one case bank (routine, 36 cases; challenging, 72 cases).
Columns are the local open-source backend (Qwen2.5), the cloud backend (GPT-5.4), and the paired difference (local minus cloud) with a 95\% confidence interval from 10,000 case-level resamples.
Field completeness is the percentage of required case fields captured; contact completeness is the pooled percentage of ground-truth contacts matched.
Interview efficiency is information units per turn; average number of interview turns is the mean conversation length; cost is LLM API spend per completed interview (\$0 for the local model).
Performance was similar across models, supporting local deployment for privacy or cloud deployment without material loss of field or contact recovery.
For local--local cost comparisons, no external API cost existed in either condition and the difference was structurally zero.
Data are mean $\pm$ SD, \%, or difference (95\% CI). CI=confidence interval, LLM=large language model, API=application programming interface, and pp=percentage points.}
\label{tab:study1_models}
\centering
\small
\begin{tabular}{lccc}
\toprule
\multicolumn{4}{l}{\textbf{Study 1 on measles: Routine bank (36 cases)}} \\
\midrule
\textbf{Metric} & \textbf{Qwen2.5 (local)} & \textbf{GPT-5.4 (cloud)} & \textbf{Difference (95\% CI)} \\
\midrule
\textbf{Field completeness (\%)} & 89.8\% & 97.1\% & $-$7.3 pp ($-$10.9 to $-$2.5) \\
\textbf{Contact completeness (\%)} & 65.3\% & 61.1\% & $+$4.2 pp ($-$11.5 to $+$20.3) \\
\textbf{Interview efficiency} & 0.605 $\pm$ 0.126 & 0.716 $\pm$ 0.150 & $-$0.112 ($-$0.152 to $-$0.072) \\
\textbf{Average Number of Interview Turns} & 39.9 $\pm$ 3.8 & 35.2 $\pm$ 4.8 & $+$4.7 ($+$3.2 to $+$6.4) \\
\textbf{Cost (\$ per completed interview)} & 0.00 & \$0.39 & $-$\$0.39 ($-$\$0.41 to $-$\$0.38) \\
\midrule
\multicolumn{4}{l}{\textbf{Study 1 on measles: Challenging bank (72 cases)}} \\
\midrule
\textbf{Metric} & \textbf{Qwen2.5 (local)} & \textbf{GPT-5.4 (cloud)} & \textbf{Difference (95\% CI)} \\
\midrule
\textbf{Field completeness (\%)} & 85.5\% & 95.3\% & $-$9.8 pp ($-$11.3 to $-$8.3) \\
\textbf{Contact completeness (\%)} & 55.5\% & 58.1\% & $-$2.7 pp ($-$12.5 to $+$7.2) \\
\textbf{Interview efficiency} & 0.545 $\pm$ 0.131 & 0.648 $\pm$ 0.158 & $-$0.103 ($-$0.125 to $-$0.082) \\
\textbf{Average Number of Interview Turns} & 39.5 $\pm$ 7.9 & 37.0 $\pm$ 7.8 & $+$2.5 ($+$1.4 to $+$3.5) \\
\textbf{Cost (\$ per completed interview)} & 0.00 & \$0.41 & $-$\$0.41 ($-$\$0.43 to $-$\$0.40) \\
\midrule
\multicolumn{4}{l}{\textbf{Study 1 on gonorrhea: Routine bank (36 cases)}} \\
\midrule
\textbf{Metric} & \textbf{Qwen2.5 (local)} & \textbf{GPT-5.4 (cloud)} & \textbf{Difference (95\% CI)} \\
\midrule
\textbf{Field completeness (\%)} & 96.4\% & 97.8\% & $-$1.4 pp ($-$2.8 to $+$0.0) \\
\textbf{Contact completeness (\%)} & 66.2\% & 69.2\% & $-$3.1 pp ($-$15.9 to $+$11.3) \\
\textbf{Interview efficiency} & 0.634 $\pm$ 0.136 & 0.558 $\pm$ 0.107 & $+$0.077 ($+$0.039 to $+$0.114) \\
\textbf{Average Number of Interview Turns} & 17.1 $\pm$ 3.5 & 19.9 $\pm$ 4.5 & $-$2.9 ($-$4.0 to $-$1.8) \\
\textbf{API Cost (\$ per completed interview)} & 0.00 & \$0.21 & $-$\$0.21 ($-$\$0.22 to $-$\$0.19) \\
\midrule
\multicolumn{4}{l}{\textbf{Study 1 on gonorrhea: Challenging bank (72 cases)}} \\
\midrule
\textbf{Metric} & \textbf{Qwen2.5 (local)} & \textbf{GPT-5.4 (cloud)} & \textbf{Difference (95\% CI)} \\
\midrule
\textbf{Field completeness (\%)} & 91.7\% & 92.9\% & $-$1.3 pp ($-$3.5 to $+$1.4) \\
\textbf{Contact completeness (\%)} & 61.7\% & 74.2\% & $-$12.5 pp ($-$22.3 to $-$2.4) \\
\textbf{Interview efficiency} & 0.570 $\pm$ 0.156 & 0.494 $\pm$ 0.130 & $+$0.076 ($+$0.050 to $+$0.104) \\
\textbf{Average Number of Interview Turns} & 18.5 $\pm$ 4.7 & 20.5 $\pm$ 5.3 & $-$2.1 ($-$2.9 to $-$1.2) \\
\textbf{API Cost (\$ per completed interview)} & 0.00 & \$0.21 & $-$\$0.21 ($-$\$0.23 to $-$\$0.20) \\
\bottomrule
\end{tabular}
\begin{flushleft}
\footnotesize\textit{Positive differences favor the local model. Absolute completeness levels should be read with the case-bank difficulty in mind: the challenging bank was written to include resistant disclosure and sparse contacts. Secondary contact-accuracy and handoff outcomes are reported in Supplementary Appendix Section 4.}
\end{flushleft}
\end{table*}

Field completeness was high with both models. For measles, the local model captured 89.8\% of required fields on the routine bank and 85.5\% on the challenging bank, versus 97.1\% and 95.3\% with the cloud model (differences $-$7.3 and $-$9.8 percentage points; both 95\% confidence intervals excluded zero). For gonorrhea, field completeness was similar across models: 96.4\% versus 97.8\% on the routine bank and 91.7\% versus 92.9\% on the challenging bank. Contact completeness was lower. The only model difference whose confidence interval excluded zero was the gonorrhea challenging bank, where the cloud model recovered more contacts (74.2\% versus 61.7\%; difference $-$12.5 percentage points, 95\% CI $-$22.3 to $-$2.4). Mean interview length for measles ranged from 39.5 to 39.9 turns with the local model versus 35.2 to 37.0 with the cloud model; for gonorrhea, means ranged from 17.1 to 18.5 turns locally versus 19.9 to 20.5 in the cloud. The local model had no API cost. Cloud cost per completed interview ranged from \$0.39 to \$0.41 for measles and was \$0.21 for gonorrhea.

The same cases were also run with each model as an open-ended interviewer without protocol rules (Table~\ref{tab:study1_openended}; Figure~\ref{fig:study_stats}A). Epira captured more required fields in every comparison. Open-ended interviewers captured 31.1 to 41.9 percentage points fewer required fields than the same models inside Epira. Open-ended interviews were much shorter: mean turns ranged from 4.2 to 15.8 for measles and from 5.4 to 9.4 for gonorrhea, versus 17.1 to 39.9 with Epira. Interview duration was also less predictable without protocol control: across conditions, the SD was 49\% to 131\% of the mean for open-ended interviews, compared with 9\% to 26\% for Epira, indicating more consistent interview length with Epira. A higher efficiency score in that arm reflects a short conversation, not a more complete case record.

\begin{table*}[!htbp]
\setlength{\abovecaptionskip}{5pt}
\caption{Primary outcomes for measles and gonorrhea (Study 1)---Epira versus the same LLMs used as fully open-ended interviewers with minimal instructions. The minimal instructions for the open-ended interviewers are provided in Supplementary Appendix Section 7.5. Without protocol guidance, both models recovered fewer fields and were less efficient, with larger gaps on the challenging bank. For local--local cost comparisons, no external API cost existed in either condition and the difference was structurally zero. Data are mean $\pm$ SD, \%, or difference (95\% CI). CI=confidence interval, LLM=large language model, API=application programming interface, and pp=percentage points.}
\label{tab:study1_openended}
\centering
\small

\begin{adjustbox}{max totalsize={\textwidth}{0.9\textheight},center}
\begin{tabular}{lccc}
\toprule
\multicolumn{4}{l}{\textbf{Measles: Routine bank (36 cases) --- Qwen2.5}} \\
\midrule
\textbf{Metric} & \textbf{Qwen2.5 (Epira)} & \textbf{Qwen2.5 (Open-ended)} & \textbf{Difference (95\% CI)} \\
\midrule
\textbf{Field completeness (\%)} & 89.8\% & 50.6\% & $+$39.1 pp ($+$31.3 to $+$46.8) \\
\textbf{Contact completeness (\%)} & 65.3\% & 49.5\% & $+$15.8 pp ($+$0.0 to $+$31.3) \\
\textbf{Interview efficiency} & 0.605 $\pm$ 0.126 & 4.328 $\pm$ 2.701 & $-$3.723 ($-$4.592 to $-$2.874) \\
\textbf{Average Number of Interview Turns} & 39.9 $\pm$ 3.8 & 4.4 $\pm$ 2.7 & $+$35.5 ($+$34.1 to $+$36.8) \\
\textbf{API Cost (\$ per completed interview)} & 0.00 & 0.00 & n/a \\
\midrule
\multicolumn{4}{l}{\textbf{Measles: Routine bank (36 cases) --- GPT-5.4}} \\
\midrule
\textbf{Metric} & \textbf{GPT-5.4 (Epira)} & \textbf{GPT-5.4 (Open-ended)} & \textbf{Difference (95\% CI)} \\
\midrule
\textbf{Field completeness (\%)} & 97.1\% & 61.9\% & $+$35.2 pp ($+$30.2 to $+$40.0) \\
\textbf{Contact completeness (\%)} & 61.1\% & 66.3\% & $-$5.3 pp ($-$20.6 to $+$9.2) \\
\textbf{Interview efficiency} & 0.716 $\pm$ 0.150 & 1.720 $\pm$ 0.863 & $-$1.003 ($-$1.265 to $-$0.737) \\
\textbf{Average Number of Interview Turns} & 35.2 $\pm$ 4.8 & 15.8 $\pm$ 18.2 & $+$19.4 ($+$12.8 to $+$24.7) \\
\textbf{API Cost (\$ per completed interview)} & \$0.39 & \$0.11 & $+$\$0.29 ($+$\$0.24 to $+$\$0.32) \\
\midrule
\multicolumn{4}{l}{\textbf{Measles: Challenging bank (72 cases) --- Qwen2.5}} \\
\midrule
\textbf{Metric} & \textbf{Qwen2.5 (Epira)} & \textbf{Qwen2.5 (Open-ended)} & \textbf{Difference (95\% CI)} \\
\midrule
\textbf{Field completeness (\%)} & 85.5\% & 54.5\% & $+$31.1 pp ($+$22.4 to $+$39.5) \\
\textbf{Contact completeness (\%)} & 55.5\% & 45.7\% & $+$9.7 pp ($-$1.7 to $+$21.6) \\
\textbf{Interview efficiency} & 0.545 $\pm$ 0.131 & 4.143 $\pm$ 3.258 & $-$3.598 ($-$4.353 to $-$2.863) \\
\textbf{Average Number of Interview Turns} & 39.5 $\pm$ 7.9 & 4.2 $\pm$ 2.6 & $+$35.3 ($+$33.2 to $+$37.0) \\
\textbf{Cost (\$ per completed interview)} & 0.00 & 0.00 & n/a \\
\midrule
\multicolumn{4}{l}{\textbf{Measles: Challenging bank (72 cases) --- GPT-5.4}} \\
\midrule
\textbf{Metric} & \textbf{GPT-5.4 (Epira)} & \textbf{GPT-5.4 (Open-ended)} & \textbf{Difference (95\% CI)} \\
\midrule
\textbf{Field completeness (\%)} & 95.3\% & 58.3\% & $+$37.0 pp ($+$32.8 to $+$40.7) \\
\textbf{Contact completeness (\%)} & 58.1\% & 61.9\% & $-$3.8 pp ($-$12.0 to $+$4.2) \\
\textbf{Interview efficiency} & 0.648 $\pm$ 0.158 & 1.768 $\pm$ 1.051 & $-$1.120 ($-$1.355 to $-$0.901) \\
\textbf{Average Number of Interview Turns} & 37.0 $\pm$ 7.8 & 10.9 $\pm$ 6.3 & $+$26.2 ($+$23.8 to $+$28.3) \\
\textbf{Cost (\$ per completed interview)} & \$0.41 & \$0.08 & $+$\$0.33 ($+$\$0.31 to $+$\$0.35) \\
\midrule
\multicolumn{4}{l}{\textbf{Gonorrhea: Routine bank (36 cases) --- Qwen2.5}} \\
\midrule
\textbf{Metric} & \textbf{Qwen2.5 (Epira)} & \textbf{Qwen2.5 (Open-ended)} & \textbf{Difference (95\% CI)} \\
\midrule
\textbf{Field completeness (\%)} & 96.4\% & 55.0\% & $+$41.4 pp ($+$33.3 to $+$49.7) \\
\textbf{Contact completeness (\%)} & 66.2\% & 47.7\% & $+$18.5 pp ($+$0.0 to $+$37.1) \\
\textbf{Interview efficiency} & 0.634 $\pm$ 0.136 & 1.528 $\pm$ 0.852 & $-$0.894 ($-$1.184 to $-$0.621) \\
\textbf{Average Number of Interview Turns} & 17.1 $\pm$ 3.5 & 5.4 $\pm$ 2.7 & $+$11.7 ($+$10.6 to $+$12.8) \\
\textbf{Cost (\$ per completed interview)} & 0.00 & 0.00 & n/a \\
\midrule
\multicolumn{4}{l}{\textbf{Gonorrhea: Routine bank (36 cases) --- GPT-5.4}} \\
\midrule
\textbf{Metric} & \textbf{GPT-5.4 (Epira)} & \textbf{GPT-5.4 (Open-ended)} & \textbf{Difference (95\% CI)} \\
\midrule
\textbf{Field completeness (\%)} & 97.8\% & 55.8\% & $+$41.9 pp ($+$38.1 to $+$45.6) \\
\textbf{Contact completeness (\%)} & 69.2\% & 67.7\% & $+$1.5 pp ($-$13.4 to $+$16.9) \\
\textbf{Interview efficiency} & 0.558 $\pm$ 0.107 & 0.944 $\pm$ 0.454 & $-$0.386 ($-$0.540 to $-$0.249) \\
\textbf{Average Number of Interview Turns} & 19.9 $\pm$ 4.5 & 9.4 $\pm$ 12.3 & $+$10.5 ($+$6.1 to $+$13.3) \\
\textbf{Cost (\$ per completed interview)} & \$0.21 & \$0.06 & $+$\$0.15 ($+$\$0.13 to $+$\$0.16) \\
\midrule
\multicolumn{4}{l}{\textbf{Gonorrhea: Challenging bank (72 cases) --- Qwen2.5}} \\
\midrule
\textbf{Metric} & \textbf{Qwen2.5 (Epira)} & \textbf{Qwen2.5 (Open-ended)} & \textbf{Difference (95\% CI)} \\
\midrule
\textbf{Field completeness (\%)} & 91.7\% & 56.1\% & $+$35.6 pp ($+$29.2 to $+$41.4) \\
\textbf{Contact completeness (\%)} & 61.7\% & 65.6\% & $-$3.9 pp ($-$16.5 to $+$8.7) \\
\textbf{Interview efficiency} & 0.570 $\pm$ 0.156 & 1.504 $\pm$ 0.984 & $-$0.934 ($-$1.157 to $-$0.727) \\
\textbf{Average Number of Interview Turns} & 18.5 $\pm$ 4.7 & 7.2 $\pm$ 8.9 & $+$11.3 ($+$8.9 to $+$13.3) \\
\textbf{Cost (\$ per completed interview)} & 0.00 & 0.00 & n/a \\
\midrule
\multicolumn{4}{l}{\textbf{Gonorrhea: Challenging bank (72 cases) --- GPT-5.4}} \\
\midrule
\textbf{Metric} & \textbf{GPT-5.4 (Epira)} & \textbf{GPT-5.4 (Open-ended)} & \textbf{Difference (95\% CI)} \\
\midrule
\textbf{Field completeness (\%)} & 92.9\% & 56.0\% & $+$36.9 pp ($+$32.4 to $+$40.7) \\
\textbf{Contact completeness (\%)} & 74.2\% & 63.3\% & $+$10.9 pp ($+$2.2 to $+$20.3) \\
\textbf{Interview efficiency} & 0.494 $\pm$ 0.130 & 0.869 $\pm$ 0.351 & $-$0.374 ($-$0.451 to $-$0.296) \\
\textbf{Average Number of Interview Turns} & 20.5 $\pm$ 5.3 & 9.1 $\pm$ 9.0 & $+$11.4 ($+$8.7 to $+$13.4) \\
\textbf{Cost (\$ per completed interview)} & \$0.21 & \$0.06 & $+$\$0.16 ($+$\$0.14 to $+$\$0.17) \\
\bottomrule
\end{tabular}
\end{adjustbox}
\end{table*}

\begin{figure*}[!htbp]
    \centering
    \setlength{\abovecaptionskip}{0pt}
    \includegraphics[width=0.8\textwidth]{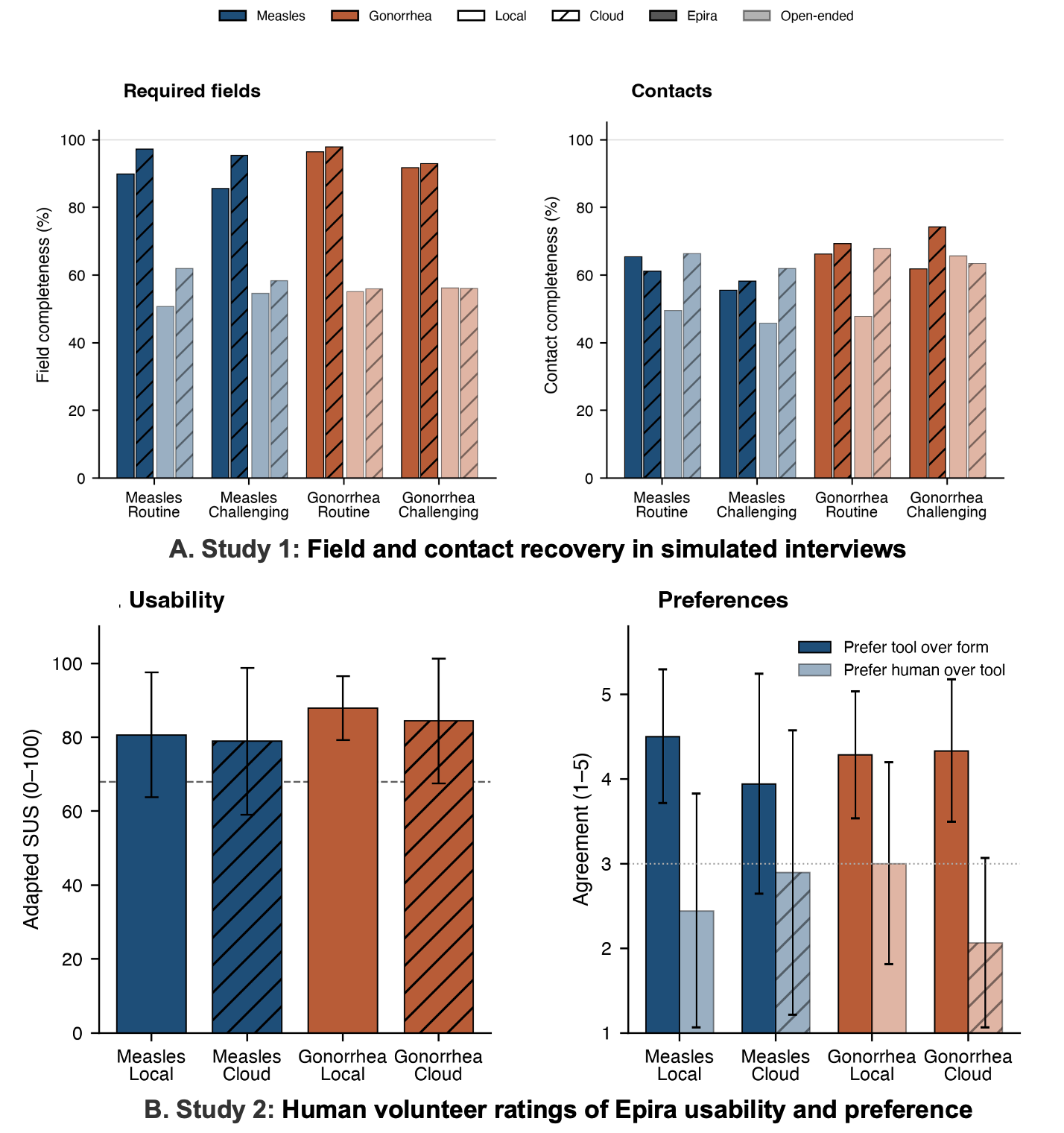}
    \caption{\textbf{Statistics for Studies 1 and 2.}
    \textbf{(A) Study 1: Field and contact recovery in simulated interviews}. Each group is one disease and one case bank (routine, 36 cases; challenging, 72 cases). The left panel shows field completeness, the percentage of that disease's required fields that were filled. The right panel shows contact completeness, the percentage of known contacts recovered. The measles banks contained 190 known contacts across 36 routine cases and 339 across 72 challenging cases; the corresponding gonorrhea banks contained 65 and 128 known contacts. \textbf{(B) Study 2: Human volunteer ratings of Epira usability and preference}. Mean $\pm$ SD scores by disease and interview engine (local Qwen2.5 vs cloud GPT-5.4); n = 18 per arm. The left panel shows the Adapted System Usability Scale (SUS; 0--100). The dashed line marks the conventional above-average SUS benchmark of 68. The right panel shows preference items (1--5): agreement with preferring the tool over a long online form, and with preferring a human interviewer over the tool. The dotted line marks the scale midpoint (3 = Neutral).}
    \label{fig:study_stats}
\end{figure*}
% \noindent{\footnotesize\textit{Alt text:} Four-panel bar charts for Study 1 field and contact completeness and Study 2 usability and preference ratings for measles and gonorrhea under local and cloud Epira deployments.}
% \FloatBarrier

\subsection*{Study 2}
Seventy-two human volunteer interviews were completed in a 2$\times$2 design (measles or gonorrhea $\times$ local Qwen2.5 or cloud GPT-5.4; n = 18 per arm). Interview completion was 100\% in every arm. Mean adapted SUS scores ranged from 78.9 to 87.9 across arms (pooled mean 83.0). The primary trust index ranged from 3.69 to 4.00 across arms, above neutral in every arm. Participants preferred the tool over a long online form (mean 4.26) more than a human interviewer over the tool (mean 2.60). Privacy comfort was near neutral (gonorrhea means 2.72 to 2.83; measles means 3.33 to 3.83). Measles interviews were more often felt to be too long (means 3.11 to 3.89) than gonorrhea interviews (means 1.61 to 1.78). Reports of technical problems were rare in every arm (means 1.06 to 1.39). 
% Across arms, comfort answering ranged from 4.11 to 4.56, and willingness to use the tool for a real case ranged from 3.72 to 4.25. 

\begin{table*}[!htbp]
\setlength{\abovecaptionskip}{5pt}
\caption{Human volunteer pilot of Epira for measles and gonorrhea (Study 2) by disease and LLM engine. Each experiment was done with 18 volunteers (72 total). User experience, acceptability and usability were assessed with a post-study questionnaire. For each metric with (1-5) scale, higher value = greater agreement. Data are mean $\pm$ SD and \%. SUS=System Usability Scale.}
\label{tab:study2}
\centering
\small
\begin{tabular}{lcc}
\toprule
\multicolumn{3}{l}{\textbf{Measles}} \\
\midrule
\textbf{Metric} & \textbf{Measles: Local Qwen2.5} & \textbf{Measles: Cloud GPT-5.4} \\
\midrule
\textbf{Interview completion rate} & 100\% & 100\% \\
\midrule
\multicolumn{3}{l}{\textbf{Usability (0-100)}} \\
\textbf{Adapted System Usability Scale (SUS)} & 80.6 $\pm$ 16.9 & 78.9 $\pm$ 19.9 \\
\midrule
\multicolumn{3}{l}{\textbf{Trust \& comfort (1-5)}} \\
\textbf{Primary trust index (mean)} & 3.91 $\pm$ 0.72 & 4.00 $\pm$ 0.84 \\
\textbf{Felt understood} & 3.83 $\pm$ 1.04 & 3.89 $\pm$ 1.13 \\
\textbf{Comfort answering} & 4.56 $\pm$ 0.62 & 4.28 $\pm$ 1.02 \\
\textbf{Privacy comfort} & 3.33 $\pm$ 1.46 & 3.83 $\pm$ 1.29 \\
\midrule
\multicolumn{3}{l}{\textbf{Preferences \& willingness (1-5)}} \\
\textbf{Willingness if real case} & 4.25 $\pm$ 0.88 & 3.89 $\pm$ 1.29 \\
\textbf{Prefer tool over form} & 4.50 $\pm$ 0.79 & 3.94 $\pm$ 1.30 \\
\textbf{Prefer human over tool} & 2.44 $\pm$ 1.38 & 2.89 $\pm$ 1.68 \\
\midrule
\multicolumn{3}{l}{\textbf{Interview burden \& problems (1-5)}} \\
\textbf{Felt too long} & 3.11 $\pm$ 1.64 & 3.89 $\pm$ 1.45 \\
\textbf{Wanted to stop for a human} & 1.89 $\pm$ 1.13 & 2.78 $\pm$ 1.63 \\
\textbf{Ran into tech problems} & 1.33 $\pm$ 0.69 & 1.33 $\pm$ 0.77 \\
\midrule
\multicolumn{3}{l}{\textbf{Process quality (1-5)}} \\
\textbf{Study clarity index (mean)} & 4.63 $\pm$ 0.50 & 4.51 $\pm$ 0.60 \\
\midrule
\multicolumn{3}{l}{\textbf{Gonorrhea}} \\
\midrule
\textbf{Metric} & \textbf{Gonorrhea: Local Qwen2.5} & \textbf{Gonorrhea: Cloud GPT-5.4} \\
\midrule
\textbf{Interview completion rate} & 100\% & 100\% \\
\midrule
\multicolumn{3}{l}{\textbf{Usability (0-100)}} \\
\textbf{Adapted System Usability Scale (SUS)} & 87.9 $\pm$ 8.6 & 84.4 $\pm$ 16.9 \\
\midrule
\multicolumn{3}{l}{\textbf{Trust \& comfort (1-5)}} \\
\textbf{Primary trust index (mean)} & 3.69 $\pm$ 0.92 & 3.80 $\pm$ 0.57 \\
\textbf{Felt understood} & 4.11 $\pm$ 1.08 & 4.22 $\pm$ 0.65 \\
\textbf{Comfort answering} & 4.11 $\pm$ 0.90 & 4.44 $\pm$ 0.70 \\
\textbf{Privacy comfort} & 2.83 $\pm$ 1.50 & 2.72 $\pm$ 1.18 \\
\midrule
\multicolumn{3}{l}{\textbf{Preferences \& willingness (1-5)}} \\
\textbf{Willingness if real case} & 3.72 $\pm$ 1.03 & 4.11 $\pm$ 0.70 \\
\textbf{Prefer tool over form} & 4.28 $\pm$ 0.75 & 4.33 $\pm$ 0.84 \\
\textbf{Prefer human over tool} & 3.00 $\pm$ 1.19 & 2.06 $\pm$ 1.00 \\
\midrule
\multicolumn{3}{l}{\textbf{Interview burden \& problems (1-5)}} \\
\textbf{Felt too long} & 1.61 $\pm$ 0.98 & 1.78 $\pm$ 1.11 \\
\textbf{Wanted to stop for a human} & 1.67 $\pm$ 0.59 & 1.61 $\pm$ 1.04 \\
\textbf{Ran into tech problems} & 1.39 $\pm$ 0.98 & 1.06 $\pm$ 0.24 \\
\midrule
\multicolumn{3}{l}{\textbf{Process quality (1-5)}} \\
\textbf{Study clarity index (mean)} & 4.58 $\pm$ 0.61 & 4.64 $\pm$ 0.66 \\
\bottomrule
\end{tabular}
\end{table*}
\FloatBarrier

\section*{Discussion}

Here, we show that a protocol-guided LLM interview system can conduct case-investigation and contact-tracing interviews for more than one disease, recover most required fields from simulated cases with known answers, and be highly rated in usability by volunteers using the system. The same interview ran on a locally hosted open-source model (Qwen2.5) and a cloud model (GPT-5.4), which yielded similar results. Local model deployment was motivated by the sensitivity of real case-investigation data.

The protocol controller (Figure~\ref{fig:epira_system}B) accounted for Epira's higher field recovery; both models recovered far more required fields when they followed Epira's protocol than when they interviewed without it. When Qwen2.5 and GPT-5.4 were used as open-ended interviewers, they captured 31 to 42 percentage points fewer required fields than the same models inside Epira (Table~\ref{tab:study1_openended}). Those open-ended interviews also ended much earlier before capturing all the required fields.

Local and cloud deployment produced similar case records. Field completeness was high with both models (Table~\ref{tab:study1_models}). The local model had no API cost. Cloud cost was about \$0.21 per completed gonorrhea interview and about \$0.40 per completed measles interview. A health department can therefore run the interview on hardware it controls with no per-interview API fee, significantly reducing costs in the case of a pandemic. Staff still review the record, take the interviews that are routed to a person, and follow the contacts the tool did not recover.

Participants in the human pilot agreed that they would rather use Epira than a long form. On average they did not need to be redirected to a human for this role-played interview (Table~\ref{tab:study2}). Epira is built to hand the session to a person when the respondent asks for one, or when the protocol flags an emergency or a situation outside the automated interview. The tool collects the structured first interview and returns a record the staff can review.

This use of a language model differs from COVID-19 self-triage and general health chatbots.\cite{sriudomporn2026,judson2020,herriman2020} Those systems checked symptoms, judged how urgent the illness was, answered questions, and managed appointments. Epira is a tool a health department can use for a first interview. It follows a disease protocol, fills a structured case record for a person to review, and elicits contacts when the protocol requires them.

Several limits apply. Simulated respondents are not patients, and the volunteers role-played fictional scenarios rather than disclosing their own contacts, so real resistance may differ. Moreover, without real index cases, we could not assess usability across sex, gender, race, ethnicity, or disability. Usability may differ across demographic groups and people with different disability-related needs. Interviews were in English and on volunteers based in the United States. Hesitant or incomplete disclosure remains the main limit of automated contact elicitation.

Future work should test Epira with real index cases, add the option of holding the interview in different languages, and evaluate the voice channel. Additional diseases could be added as ready-to-use standard protocols. In future usability testing with real index cases, we will analyze whether usability differs by sex, gender, race, ethnicity, or disability.

\section*{Conclusion}
We presented Epira, an automated, protocol-guided conversational interviewer for case investigation and contact tracing, designed to support multiple diseases on one shared interview workflow, together with a two-part evaluation plan using simulated index cases and human volunteers. Experiments showed that Epira can recover most required case fields and contacts. Volunteer users found it easy to use and preferred it over filling out forms.

\section*{Acknowledgments}
The authors gratefully acknowledge support from the Office of Naval Research (ONR) award N000142412306, Air Force Office of Scientific Research (AFOSR) award FA9550-26-1-0012, the Alfred P. Sloan Foundation, the Stanford Institute for Human-Centered Artificial Intelligence (HAI), and from IBM Research as a founding member of Stanford Institute for Human-centered Artificial Intelligence. 

\section*{Data Sharing}
Supplementary Appendix Sections 1--5 and 7 report the disease specifications, controller rules, interviewer prompts, Study 2 questionnaire, and one full persona from each case bank. Study 1 used synthetic cases. No individual participant data will be shared as Study 2 used fictional scenarios only and no patient data were collected. The Epira website is available for review at \url{https://contact-tracing-demo.onrender.com/}. The source code is not publicly released. The disease specifications, controller rules, and interviewer prompts used in the reported runs appear in Supplementary Appendix Sections 1--3 and 7.

\section*{Conflicts of interest}
The authors declare no competing interests.

\section*{Generative AI statement}
During preparation of this manuscript, the authors used GPT-5.6 Sol for language editing and grammatical review. The authors reviewed and edited all resulting text and take full responsibility for the content of the manuscript. In this work, Qwen2.5 and GPT-5.4 were used as interview LLM engines and are described in the Methods. GPT-5.4-mini was used to simulate users in Study 1.

\bibliographystyle{vancouver}
{\footnotesize \bibliography{citation}}

\end{document}